\newif\ifisr
\isrtrue

\ifisr
    \documentclass[orange]{isrpaper}
    \newcommand{\IEEEPARstart}[2]{#1#2}
    \newenvironment{IEEEkeywords}%
        {\par\vspace{4pt}\noindent\small\textbf{Keywords: }}%
        {\par\normalsize\vspace{6pt}}
    \renewcommand{\thanks}[1]{}
    \usepackage[
        backend=biber,
        style=authoryear,
        natbib=true,
        doi=false,
        url=false,
        isbn=false,
        maxbibnames=99,
        maxcitenames=2,
        uniquelist=false,
    ]{biblatex}
    \usepackage{orcidlink}
    \renewcommand{\cite}[1]{\parencite{#1}}
\else
    \documentclass[11pt]{article}
    \usepackage[margin=1in]{geometry}
    \usepackage{hyperref}
    \hypersetup{hidelinks=true}
\fi

\usepackage{diagbox}
\usepackage{amsmath,amssymb,amsfonts}
\usepackage{graphicx}
\usepackage{algorithm,algorithmic}
\usepackage{textcomp}
\usepackage{booktabs}
\usepackage{overpic}
\usepackage{colortbl}
\usepackage[font=footnotesize]{subfig}
\usepackage{threeparttable}
\usepackage{placeins}  

\newcommand{\biomechgptabstract}{%
Advances in markerless motion capture are making high-quality biomechanical data increasingly accessible, creating a growing need for scalable downstream analytics. Building a bespoke pipeline for each analysis task is time-consuming, motivating models that can flexibly handle diverse clinical questions within a single framework. Recent work has shown that fine-tuning language models to accept tokenized motion as an additional modality enables descriptive captioning of movement, raising the question of whether these models are also capable of clinically relevant motion understanding, where diverse tasks and annotations provide a natural testbed. We investigate whether such a multimodal motion--language model can answer detailed, clinically meaningful questions about movement. We collected 71 hours of biomechanical data from 750 participants, many with movement impairments, performing tasks commonly used in clinical assessment. To further expand the training dataset, we designed a cross-format tokenizer that directly encodes motion data from heterogeneous formats into a shared latent space without paired data, allowing a second dataset to be incorporated and enabling pooling annotations across datasets. From these tokenized representations, we constructed a multimodal dataset of motion-related question--answer pairs and used it to train BiomechGPT, a multimodal biomechanics--language model. BiomechGPT achieves competitive performance across a range of clinically relevant tasks, with performance scaling with both dataset and model size. It offers a new way for clinicians and researchers to interact with biomechanical data and represents a promising direction for rehabilitation-focused movement analysis. Project page: \url{https://intelligentsensingandrehabilitation.github.io/BiomechGPT/}%
}

\begin{document}
\ifisr
    \title{BiomechGPT: Extending Motion-Language Models to Clinical Motion Understanding}
    \author{Ruize Yang\orcidlink{0009-0006-8353-165X}$^{1, 2}$, Ann Kennedy\orcidlink{0000-0002-3782-0518}$^{3}$, R. James Cotton\orcidlink{0000-0001-5714-1400}$^{1, 4}$}
    \affiliations{%
        $^{1}$Center for Bionic Medicine, Shirley Ryan AbilityLab, Chicago, IL \quad
        $^{2}$Department of Neuroscience, Northwestern University, Chicago, IL \quad
        $^{3}$Department of Neuroscience, The Scripps Research Institute, San Deigo, CA \quad
        $^{4}$Department of Physical Medicine \& Rehabilitation, Northwestern University, Chicago, IL%
    }
    \correspondence{rcotton@sralab.org}
    \paperabstract{\biomechgptabstract}
    \maketitle
\fi

\begin{figure*}[!t]
    \centering
    \includegraphics[width=0.9\linewidth]{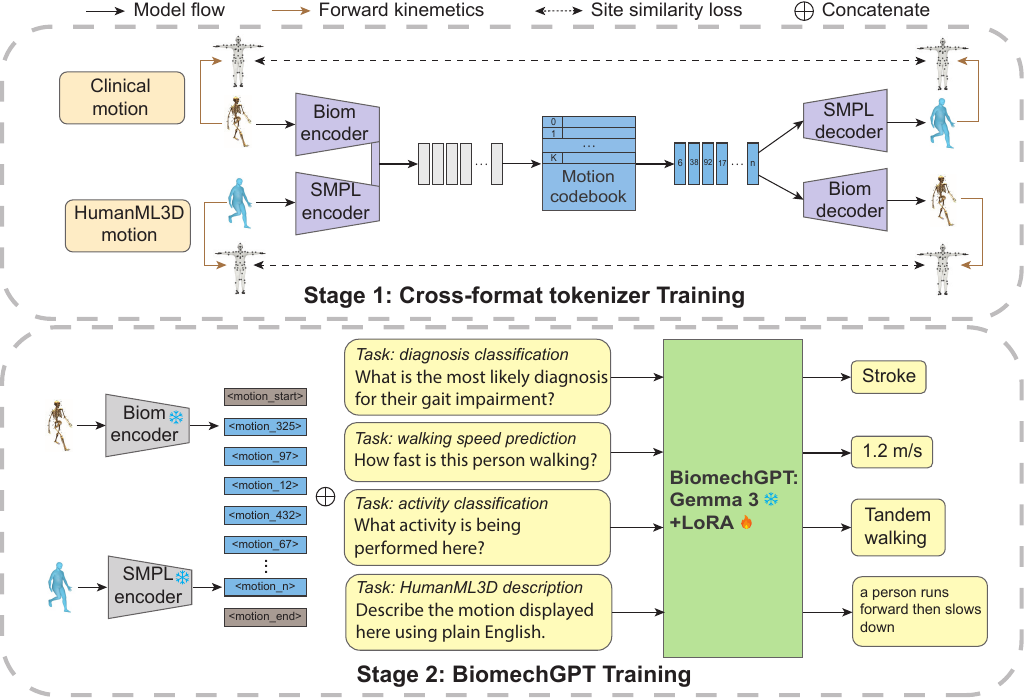}
    \caption{\textbf{Overview of BiomechGPT. }
    \textbf{Stage 1: Cross-format tokenizer training.} We trained a VQ-VAE-based tokenizer jointly on the Clinical dataset (in biomechanical model (Biom) format) and the HumanML3D dataset (in SMPL format). Format-specific encoders embed motion from either format into a shared motion codebook, and format-specific decoders reconstruct motion in both formats from the quantized tokens. A site-similarity loss compares site locations between the input motion and the cross-format decoded output (obtained via forward kinematics), enforcing the shared latent space without requiring paired samples across formats.
    \textbf{Stage 2: BiomechGPT training.} We froze the tokenizer and used it to convert each motion trajectory into a sequence of discrete tokens. We then concatenated the motion tokens with a natural-language question prompt, and passed this to a Gemma 3 language model to fine-tune with QLoRA, producing a single model (BiomechGPT) trained to answer diverse clinical questions. (Snowflake and flame icons indicate frozen and trainable components, respectively.)}
    \label{fig:model_structure}
\end{figure*}

\section{Introduction}
\label{sec:introduction}
\IEEEPARstart{H}{ow} someone moves carries rich information about their health \cite{Baker_2006_Gait_Analysis}. Recent advances in markerless motion capture have made it substantially easier to capture biomechanics in clinically accessible settings \cite{kanko_2021_concurrent, uhlrich_2023_opencap}. Movement can now be recorded in several ways, ranging from multi-camera systems \cite{cotton_differentiable_2025} to a single smartphone camera \cite{peiffer_2025_pbl}. While this accessibility lets clinicians and scientists measure more aspects of movement, a remaining barrier to clinical impact is analyzing the resulting data. People perform a nearly unlimited range of movements, and the questions one might ask range from measuring specific biomechanical parameters to high-level tasks such as producing a differential diagnosis or scoring a clinical outcome assessment. A substantial body of work has applied machine learning to such clinical gait tasks, including disease classification \cite{Caramia_2018_PD_cls, Figueiredo_2018_ml_gait_analysis}, fall-risk prediction \cite{howcroft_2017_fall_pred}, and gait pattern identification \cite{horst_2019_explaining, Halilaj_2018_ml_Biomechanics}, but each pipeline is typically engineered for a single question. It becomes infeasible to develop bespoke analysis pipeline for each task as accessible motion analysis becomes more widely used. Large language models (LLMs) offer a natural framework for unifying these distinct tasks under a single model \cite{raffel_2023_t5} and supporting flexible, language-based interaction with movement data by clinicians.

Building on recent advances in multi-modal LLMs \cite{liu_2023_llava, Xiao_2025_medical_llm}, a new class of motion-language models has begun to emerge. These models first train a vector-quantized variational autoencoder (VQ-VAE) \cite{oord_2018_vqvae} on motion sequences to convert them into discrete tokens, then extend the language model's vocabulary to include this new modality and train it to both consume and produce motion tokens, enabling the model to caption and generate motion. This tokenizer-based architecture was originally developed for text-conditioned human motion generation \cite{zhang_t2mgpt_2023}, where the autoencoder structure naturally supports synthesis and where the available datasets typically pair motion with natural language descriptions \cite{guo_generating_2022}. It has since been widely adopted in works that treat motion as an additional modality for language models \cite{jiang_motiongpt_2023, zhou_2023_avatargpt, jiang_2024_motionchain, luo_2024_m3gpt, chen_language_2024, ling_2025_VersatileMotion, li_2025_lamp}, with a focus on bidirectional generation between motion and its textual description. Within this paradigm, motion understanding (i.e., description) is most often coupled with motion generation \cite{chen_2024_motionllm_movid} and evaluated through motion captioning using indirect metrics such as linguistic scores or motion–text feature similarity \cite{guo_generating_2022, guo_2022_tm2t}; recently, a handful of works have begun to probe motion understanding more directly via conversation \cite{chen_2024_motionllm_movid} and question answering \cite{zhang_2026_smd}. This motivates us to ask whether the motion-language model framework can be extended to clinical motion analysis, a domain that offers diverse tasks, precise ground truths, and a focus on motion understanding that carries direct clinical meaning. We explore this direction by training a multimodal motion-language model that jointly models human biomechanics and natural language, which we call BiomechGPT.

One major challenge in adapting general-purpose motion models to clinical analysis lies in differences in the underlying body models used across the human motion datasets, which use various skeletons and surface markers. Prior works \cite{guo_generating_2022, lin_2024_motionx} mainly represent motion with the SMPL family of models \cite{SMPL:2015, pavlakos_2019_smplx}; however, these models have a kinematic chain that is not biomechanically accurate \cite{Keller_2023_skel} and does not reflect how clinicians describe movement. Clinical motion analysis instead parameterizes movement using biomechanical models, such as those used in \cite{caggiano_2022_myosuite, alhafez_2023_locomujoco}. Motivated by \cite{sárándi_2022_metrabs}, instead of converting multiple motion datasets with different skeleton and joint definitions into a single canonical skeleton, we develop a cross-format VQ-VAE tokenizer that embeds motion data in different representations directly into a shared latent space, without requiring paired samples across formats. This approach leverages the body-model-based structure of human motion data and can be extended to additional datasets as long as similarity metrics between  marker/joint positions across body models can be defined. It efficiently expands available training data, enables annotation sharing and format conversion across datasets, and makes it possible to study scaling effects despite the relative scarcity of clinical motion data.

To evaluate our framework on real clinical tasks, we collected a large dataset comprising 71 hours of biomechanics data from 750 participants, many with movement impairments from various etiologies such as lower-limb prosthesis use or neurological injury. Our data was collected using either multi-camera system or single smartphone video, letting us test whether our method works with different data acquisition systems. Participants performed a wide range of activities, with an emphasis on standardized clinical outcome assessments for mobility. The dataset is annotated with activity type, gait impairment, impairment etiology, assistive device use, clinical assessment performance, and fall history (Table~\ref{table:dataset_size}). We define a set of clinically meaningful tasks in natural language form, which enables us to test whether motion–language models can solve fine-grained clinical tasks and generalize to populations with movement impairments. This contrasts with prior motion language models that produce open ended descriptions. Our results show that a multimodal motion-language model can solve these detailed tasks on clinical biomechanics data acquired from either multi-camera recordings or smart phone videos, with performance improving with larger model size and more training data (Fig~\ref{fig:model_scaling}). Together, these findings suggest a promising role for motion-language models in clinical motion science and rehabilitation.

In summary, our contributions are:
\begin{itemize}
\item We curate a large biomechanics–language dataset for clinically relevant movement understanding, covering activity recognition, impairment and diagnosis identification, and clinical scoring. It serves as a benchmark of clinically meaningful motion understanding ability for motion LLMs.
\item We develop a cross-format VQ-VAE tokenizer that directly embeds human motion data in different formats into a shared latent space. This format-invariant representation of motion enables cross-dataset annotation-sharing, format conversion, and studying scaling effects, with the potential to be extended to additional datasets.
\item We train BiomechGPT, a multimodal motion–language model that performs well across diverse clinically meaningful question–answer tasks, on data captured from either multi-camera systems or a single smartphone. We characterize per-task performance and compare with a non-language-based model, and show performance gains from scaling both dataset size and model size. BiomechGPT offers a new way for clinicians and researchers to interact with movement analysis data using natural language.
\end{itemize}

\section{Methods}

\subsection{Datasets}
We used two datasets with distinct pose representations in training our model: a clinical dataset collected in-house, and the public HumanML3D dataset \cite{guo_generating_2022}. Both datasets capture the 3D pose and appearance of human participants, however the means of acquisition and the format of pose representations differ between them, and there is no paired sample across formats.

In constructing the input to our motion understanding model, we specifically exclude body shape and root (pelvis) location signals from both datasets, to ensure that the representation is shape and location invariant. We represent pose data purely in terms of joint angles in the input to our model; as joint angles are the how biomechanists and clinicians characterize movement \cite{wu_isb_2002,wu_isb_2005}. These joint rotations also serve as the input pose parameter for the corresponding body model, allowing our tokenizer to use the model-based property of motion datasets.

\subsubsection{Clinical dataset}
We collected a clinical motion dataset comprising 71.1 hours of biomechanical trajectories from 750 participants performing a range of activities, many of which are standard clinical outcome assessments for mobility (e.g., overground walking, the timed-up-and-go test (TUG), the L-test, the four-square-step test (FSST), and functional gait assessments). Many participants had gait impairments, including lower-limb amputation with prosthetic use, as well as neurological conditions such as stroke. Participants were recorded with single camera or multiple camera systems, and the videos were processed with markerless motion capture algorithms to generate kinematic trajectories. We processed our multiview markerless motion capture data following the approach of~\cite{cotton_differentiable_2025}, using a biomechanical model modified from LocoMujoco~\cite{alhafez_2023_locomujoco}. Monocular video collected with a smartphone was processed with our Portable Biomechanics Laboratory system \cite{peiffer_2025_pbl}. All data collection was approved by Northwestern University Institutional Review Board (STU00215263), informed consent was received from all participants. Details of the biomechanical reconstruction are provided in the supplementary material (suppl. section~\ref{suppl:biom_method}).

Each trial in our dataset is a trajectory $\boldsymbol{\tau} = \{\boldsymbol{\beta}, \mathbf{q}_0, \mathbf{q}_1, \ldots, \mathbf{q}_T\}$, where $\boldsymbol{\beta}$ is body shape and $\mathbf{q}$ is a 41-dimensional vector in which the first 3 elements give the pelvis position in Euclidean space, the next 4 give a quaternion representation of pelvis orientation, and the remaining 34 elements give body joint angles, each corresponding to a defined degree of freedom. To eliminate variance unrelated to movement content, we discard body shape and pelvis position and orientation. We also zero out wrist flexion, deviation, and supination, as well as the metatarsal joint of the foot, since these are tracked less reliably in our markerless motion capture system. We use an 80/20 train/test split at the participant level, kept consistent across both tokenizer and BiomechGPT training. Table~\ref{table:dataset_size} summarizes the split.

\subsubsection{HumanML3D dataset.}
As a second dataset, we use the publicly available HumanML3D dataset~\cite{guo_generating_2022}, which contains motion samples of participants without movement impairments performing daily activities paired with natural language descriptions. HumanML3D is largely derived from the AMASS dataset~\cite{AMASS:ICCV:2019} in SMPL format and is widely used for motion modeling. It represents motion with an overcomplete 263-dimensional vector that includes joint locations, rotations, velocities, and binary contact features. However, this representation cannot be converted back to SMPL format, as it discards some pelvis information and is retargeted to a default skeleton. To use the SMPL body model and stay consistent with the processing of our clinical dataset, we instead use the original SMPL data corresponding to each HumanML3D sample and upsample them to 30Hz, then mirroring left and right following original processing steps. We restrict this to the samples available in SMPL format, i.e. the part of HumanML3D derived from AMASS rather than from HumanAct12~\cite{Guo_2020_humanact12}. From these, we take the SMPL pose parameters (3-D joint rotations) for 21 body joints and discard global translation, body shape, and pelvis rotation. Notably, the SMPL format differs from biomechanical models in its skeleton, joint definitions, and body parameterization \cite{Keller_2023_skel, keller_2022_osso}, motivating the need for cross-dataset alignment.

\subsection{Cross-format Motion Tokenizer}
Our motion tokenizer is based on the VQVAE architecture used in \cite{jiang_motiongpt_2023, zhang_t2mgpt_2023}, with 1D-convolutional encoders and decoders that convert joint angle trajectories into sequences of discrete tokens. These tokens are drawn from a codebook of $K$ vectors of dimension $D$, learned during training. The encoder downsamples an input sequence of length $L$ by a factor of $l$, producing a latent sequence of length $L/l$. At each timestep, the latent sequence is replaced by its nearest codebook vector under Euclidean distance, and a symmetric 1D-convolutional decoder reconstructs the motion from the quantized embeddings. A commitment loss pulls the encoder output $z_e$ toward its corresponding codebook vector $e$:
\begin{equation}
\mathcal{L}_{\mathrm{commit}} = \left\lVert z_e - \mathrm{sg}[e] \right\rVert_2^2,
\label{eq:loss_commit}
\end{equation}
where $\mathrm{sg}(\cdot)$ denotes the stop-gradient operator. Following common practice, the codebook is updated via exponential moving average and codebook resetting to encourage convergence and high codebook utilization~\cite{oord_vqvae_2018, williams_hierarchical_2020}.

We further designed our tokenizer to embed different motion data formats (here, biomechanical or HumanML3D/SMPL motion data) into a shared, format-agnostic latent space (Fig.~\ref{fig:model_structure}). Datasets that exist in more than one motion format are rare, thus we developed a method to learn a shared representation without using paired data. Specifically, we measured the difference between a biomechanical pose and a SMPL pose by matching marker sites from the biomechanical model to vertices from the SMPL model. Our biomechanical model uses the same 87 marker sites as the BML-MoVi dataset~\cite{ghorbani_movi_2021}; among these, we identified 56 body-surface sites that have one-to-one correspondences with SMPL mesh vertices, obtained from~\cite{ghorbani_soma_2021, pavlakos_smplx_2019}. Since no mapping between MoVi and SMPL exists for the head, we define three additional sites (jaw, top of head, and back of head) on SMPL vertices and added markers at the corresponding locations on our biomechanical model, yielding 59 sites in total.

To support tokenization of pose data in either format, our tokenizer has two format-specific encoders that share a final layer, and two decoders that map from the shared latent space back to their respective formats. Each training sample is in single-format, i.e., either biomechanical or SMPL. During training, a given motion sample (e.g. $X_{\mathrm{bio}}$ in biomechanical format) is embedded using its format-specific encoder to produce a token sequence; that sequence is then fed into both decoders: the biomechanical decoder reconstructs the input as $\hat{X}_{\mathrm{bio}}$, while the SMPL decoder converts the same motion content to SMPL format as $\hat{X}_{\mathrm{smpl}}$. The output of these decoders is used in two loss terms: a reconstruction loss
\begin{equation}
\mathcal{L}_{\mathrm{recon}} = \mathrm{SmoothL1}(X_{\mathrm{bio}}, \hat{X}_{\mathrm{bio}}),
\end{equation}
and cross-format site similarity loss. For the latter, we compare the ground-truth biomechanical site locations with the matched vertices obtained from $\hat{X}_{\mathrm{smpl}}$. Using the forward kinematics (FK) operators $FK_{\mathrm{bio}}(\cdot)$ and $FK_{\mathrm{smpl}}(\cdot)$, we have
\begin{equation}
\begin{aligned}
\mathrm{site}_{\mathrm{bio}} &= \mathrm{sg}\!\left(FK_{\mathrm{bio}}(X_{\mathrm{bio}})\right),\\
\mathrm{site}_{\mathrm{smpl}} &= FK_{\mathrm{smpl}}(\hat{X}_{\mathrm{smpl}}).
\end{aligned}
\end{equation}
The site similarity loss is then
\begin{equation}
\mathcal{L}_{\mathrm{site\_sim}} = \frac{1}{3N} \sum \left\lVert \mathrm{site}_{\mathrm{bio}} - \mathrm{site}_{\mathrm{smpl}} \right\rVert_2^2,
\end{equation}
where $N = 59$ and $\mathrm{site}_{\mathrm{bio}}, \mathrm{site}_{\mathrm{smpl}} \in \mathbb{R}^{N \times 3}$. Both FK operators are differentiable and kept frozen. The process is symmetric for SMPL inputs, with the loss terms computed analogously.

Since we discard root information, we feed zero root location and fixed root orientation to the FK operators to align the output orientations of the two formats. The model also learns a global 10-D body shape parameter and a global 3-D location offset for SMPL across the dataset, accounting for differences in default body shape and root location between the two body models. $\mathcal{L}_{\mathrm{smpl\_shape}}$ is an $L_2$ regularization on the learned SMPL shape parameter.

Finally, the SMPL format represents pose as 3D rotations of the joints, which gives it more degrees of freedom than actual human skeleton, thus many different SMPL configurations can give the same pose, including implausible ones. We therefore constrain rotation to be minimal by applying an $L_1$ regularization to the learned SMPL pose:
\begin{equation}
\mathcal{L}_{\mathrm{smpl\_pose}} = \left\lVert \hat{X}_{\mathrm{smpl}} \right\rVert_1.
\end{equation}

Training minimizes a combination of these loss terms:

\begin{equation}
\begin{aligned}
\mathcal{L} =\; & {\color{blue} \underbrace{\mathcal{L}_{\mathrm{recon}} + \beta_{\mathrm{commit}} \mathcal{L}_{\mathrm{commit}}}_{\text{VQ-VAE Objectives}}} \\
&+ {\color{teal} \underbrace{\mathcal{L}_{\mathrm{site\_sim}}}_{\text{Cross-Format Alignment}}} \\
&+ {\color{red}  \underbrace{\beta_{\mathrm{pose}} \mathcal{L}_{\mathrm{smpl\_pose}} + \beta_{\mathrm{shape}} \mathcal{L}_{\mathrm{smpl\_shape}}}_{\text{SMPL Regularization}}}
\end{aligned}
\end{equation}

Following existing work\cite{jiang_motiongpt_2023}, we use a temporal downsampling factor of $4$ from raw pose trajectories to tokens, codebook size $K = 512$, codebook dimension $D = 512$, and $\beta_{\mathrm{commit}} = 0.02$. $\beta_{\mathrm{shape}} = 0.001$ and $\beta_{\mathrm{pose}} = 0.001$ were selected empirically. Tokenizer training takes about 20 hours on a single A6000 GPU with a batch size of 256; each batch mixed clinical and HumanML3D data 50/50. 

During training, we found that the reconstruction loss term on the clinical data validation set converged earlier than that on HumanML3D data validation set. As our target downstream tasks are focused on the clinical dataset, we selected the checkpoint where its reconstruction loss converged as the model to use for tokenization and training BiomechGPT. During tokenization, this frozen model is used to convert each pose trajectory into a sequence of motion tokens.

\subsection{Clinical Annotation and Multimodal Dataset Creation}
We next created a collection of motion assessment tasks to evaluate model performance on. For each task, we constructed a set of multimodal (text + motion tokens) prompts, which take the form of question–answer pairs such as:

\texttt{Q: <motion\_start> <motion\_1> \ldots <motion\_N> <motion\_end> What is the most likely \\ diagnosis for this gait impairment? A: Stroke.}

Table~\ref{table:dataset_size} shows the train/test split for each task. For each task we created a set of question prompts; prompts are sampled randomly during training and fixed at test time. The full list of tasks and example prompts is provided in the supplementary material (suppl. section~\ref{suppl:all_qa}).

\subsubsection{Classification tasks} We created six classification tasks from ground-truth labels in our clinical dataset, including the activity being performed, whether the participant had a gait impairment, the impairment etiology, assistive device presence and type, and, for prosthesis users, their fall history (whether they had fallen in the past 12 months). We generated training samples for all applicable tasks; for example, we did not create a diagnosis sample for participants without a gait impairment, nor a fall-history sample when that information was not collected.

\subsubsection{Regression tasks} Some clinical movement assessments have continuous-valued measurements, such as the time to complete a Timed-Up-and-Go (TUG) or Four-Square-Step Test (FSST). In addition, some trials in our clinical dataset included participants walking on an instrumented walkway (GAITRite) that measured their walking speed and cadence. We thus created additional prompts that ask the model to predict TUG and FSST time, walking speed, or walking cadence for applicable trials. Numerical answers were rounded to one decimal place, or to the nearest integer for walking cadence.

\subsubsection{HumanML3D tasks} Finally, we created an additional set of tasks using the annotations available with the original HumanML3D dataset. The first asks the model to generate a natural-language description of a given motion sequence, using the text annotations provided in HumanML3D. For trials with partial-sequence descriptions and start/end timestamps, we also created action localization tasks in which the model is asked to predict the start time, end time, duration, or description of the action during that time window.

\begin{table}[ht]
    \centering
    \footnotesize
    \setlength{\tabcolsep}{3pt}
    \caption{Size of train/test split by number of samples and participants.}
    \begin{tabular}{lcccc}
    \toprule
    Task & \shortstack{Train \\ Participants} & \shortstack{Train \\ Samples} & \shortstack{Test \\ Participants} & \shortstack{Test \\ Samples} \\
    \midrule
    \multicolumn{5}{l}{\textit{Clinical dataset: 71.1 hours, 12954 trials}} \\
    \midrule
    \multicolumn{5}{l}{\textbf{Classification Tasks}} \\
    \midrule
    activity classification & 597 & 9967 & 151 & 2433 \\
    impaired classification & 393 & 7916 & 97 & 1802 \\
    diagnosis classification & 218 & 3958 & 57 & 869 \\
    assistive device presence & 483 & 8373 & 127 & 2178 \\
    assistive device type & 104 & 994 & 17 & 138 \\
    fall history classification & 33 & 792 & 11 & 226 \\
    \midrule
    \multicolumn{5}{l}{\textbf{Regression Tasks}} \\
    \midrule    
    gaitrite cadence & 109 & 1236 & 27 & 243 \\
    gaitrite walking speed & 109 & 1236 & 27 & 243 \\
    TUG time prediction & 131 & 361 & 37 & 101 \\
    FSST time prediction & 80 & 245 & 23 & 71 \\
    \midrule
    Total participant count & 598 & - & 152 & - \\
    \midrule
    \multicolumn{5}{l}{\textit{HumanML3D dataset: 27.5 hours, 13423 trials}} \\
    \midrule
    description & - & 66994 & - & 11882 \\
    range time & - & 1820 & - & 332 \\
    range duration & - & 1820 & - & 332 \\
    range description & - & 1820 & - & 332 \\
    \bottomrule
    \end{tabular}
    \label{table:dataset_size}
\end{table}

\subsection{Multimodal Finetuning}
Having established a set of multimodal motion-language tasks, we next fine-tuned a series of language models on them. Before training, we extended the language model vocabulary to include the motion tokens. Our training protocol follows that of MotionLLM~\cite{wu_motionllm_2024}. As our language model backbone, we use pre-trained Gemma~3~\cite{gemmateam_gemma3_2025} models with 1B, 4B, or 12B parameters, as well as MedGemma-1.5~\cite{sellergren_medgemma_2025} with 4B parameters. We performed supervised fine-tuning with a quantized low-rank adapter (QLoRA)~\cite{dettmers_qlora_2023}, implemented in the HuggingFace Transformers library~\cite{wolf_huggingface_2020} with 4-bit quantization. We use a QLoRA rank and alpha of 32, a dropout of 0.1, a learning rate of $2 \times 10^{-4}$, and a maximum sequence length of 1024. The dataset was shuffled to interleave tasks, and we trained for 1.2 epochs, as preliminary experiments showed overfitting beyond a single epoch.

We trained single models across all tasks using the instruction-tuned base model with the standard Gemma chat template. Each question from our dataset was inserted as a user turn, with the answer as the model output, and we used single-turn supervised fine-tuning. Because our prompts contain the newly added motion tokens but are otherwise short, we included all prompt tokens in the loss computation, which is known to improve performance \cite{shi_2024_instruction_loss} (see also Table~\ref{table:tokenization_effect} row 1 vs 4). Training BiomechGPT on the 10 clinical tasks took between 1 and 7 hours on a single A6000 GPU, depending on the backbone size.

\subsection{Transformer Baseline}
As a point of comparison for the language models, we trained a set of non-language transformer models~\cite{vaswani_transformer_2023} on the 10 tasks from the clinical dataset. Each input sequence was prepended with a task-specific special token, and task-specific linear heads attached after the final transformer layer were trained to produce the target output from the representation of this token. The model was trained using cross-entropy loss for classification tasks and SmoothL1 loss for regression tasks; classification loss was normalized by $\log N$ (where $N$ is the number of class labels), and regression target values were standardized to zero mean and unit variance. Other aspects of dataset processing were kept the same as for BiomechGPT. The encoder has 6 layers, 4 attention heads, a model dimension of 256, and a dropout rate of 0.2.

To quantify effects of tokenization on performance, we trained two variants of this model: the transformer-token model takes motion tokens as input and is directly comparable to the language model backbone, while the transformer-raw model takes raw joint angles passed through a linear projection layer, receiving continuous representations. The transformer-token model was trained with a maximum sequence length of 1024 and a learning rate of $2 \times 10^{-4}$, and took 7 hours to train on a single A6000 GPU. The transformer-raw model used a maximum sequence length of 4096 (as we downsampled pose data by a factor of 4 in generating motion tokens) and a learning rate of $5 \times 10^{-5}$, and took 40 hours to train on the same hardware.

\subsection{Tokenizer Variation Experiments}
To test the effect of tokenizer latent capacity on downstream task performance (for background information, see section~\ref{subsec:tokenization}) within our BiomechGPT framework, we conducted a series of studies using the 4B Gemma-3 backbone and the 10 clinical task set. First, to test whether larger codebook capacity improved performance on downstream tasks, we compared the default VQVAE ($K=512$) against a larger VQVAE ($K=2048$) and then against a residual VQVAE~\cite{guo_2023_momask} with two $K=256$ codebooks.

Second, because VQVAE-based motion-language models typically discard the learned VQVAE codebook and learn the embeddings for the motion tokens from scratch in the downstream model, we explored soft tokens as an alternative. Soft tokens were obtained either from the learned VQVAE codebook embeddings or by replacing the VQVAE tokenizer with a regular VAE, in which the commitment loss (Eq.~\ref{eq:loss_commit}) is replaced by a standard KL loss. Hard tokens are the discrete codebook indices fed to the language model (same as other experiments and prior works). Soft tokens are continuous embeddings (learned codebook entries for the VQVAE, latents for the VAE) and a two-layer MLP projects soft tokens into the language model's embedding space to match the dimension. Following prior motion-tokenizer work, both the VQVAE and VAE were trained on fixed-length windows and applied to full trials at inference. The VQVAE generalized well to longer inputs, whereas the VAE reconstruction quality degraded substantially (suppl.\ section~\ref{suppl:input_len}, similar degradation was also reported in \cite{petrovich_2021_actor}). This let us obtain soft tokens with different reconstruction quality from the same VAE model, by encoding the full trial either in a single pass or as concatenated windows, enabling an additional comparison of embedding quality. Because cross-entropy does not apply to continuous targets, soft-token models compute the loss only over the answer portion; for a matched comparison, we retrained the hard-token VQVAE with the same answer-only loss. The soft-token MLP and the hard-token embedding layer were each pretrained for one epoch in this experiment.

\section{Results}
We first present the training of our motion tokenizer, demonstrating that two distinct pose formats can be successfully embedded into a shared latent space (Section~\ref{subsec:tokenizer}). We next evaluate BiomechGPT across a range of tasks, showing example results on classification (Section~\ref{subsec:cls_task}) and regression (Section~\ref{subsec:reg_task}) tasks, and the model's instruction following ability (Section~\ref{subsec:instruct_follow}). We also show the model works with biomechanics data collected in different settings (Section~\ref{subsec: pbl}). We then present the scaling behavior from increasing language model size and training dataset size (Section~\ref{subsubsec: scaling}), and compare BiomechGPT to non-language motion transformers (Section~\ref{subsec:baseline_compare}). Finally, we use our clinical task framework to characterize the effect of different tokenizer choices on task performance (Section~\ref{subsec:tokenization}).

Numbers in the tables are averaged across three independent repeats, while figures display results from one example run unless otherwise specified.

\subsection{Tokenizer Training}
\label{subsec:tokenizer}
\begin{figure*}[!t]
    \centering
    \begin{minipage}[]{\linewidth}
        \centering
        \begin{minipage}[]{0.37\linewidth}
            \centering
            \vspace{10pt}
            \begin{overpic}[width=\linewidth]{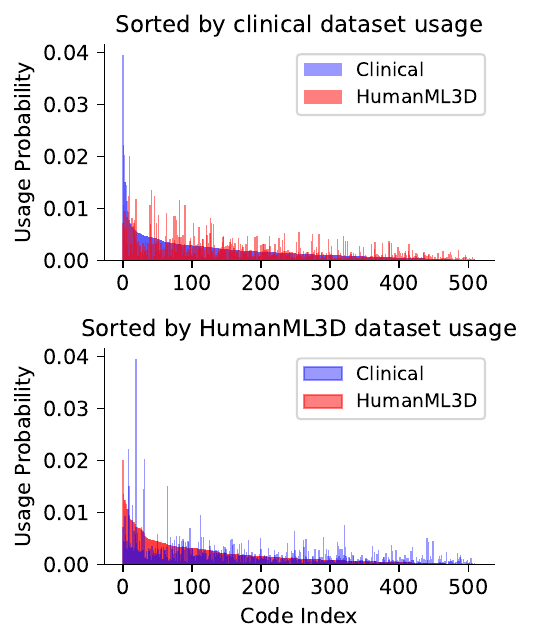}
                \put(2,99){\textbf{(a)}}
            \end{overpic}
        \end{minipage}
        \hfill
        \begin{minipage}[]{0.6\linewidth}
            \textbf{(b)}\\[2pt]
        \begin{tabular}{@{}cc@{}}
            \fbox{\includegraphics[width=0.47\linewidth]{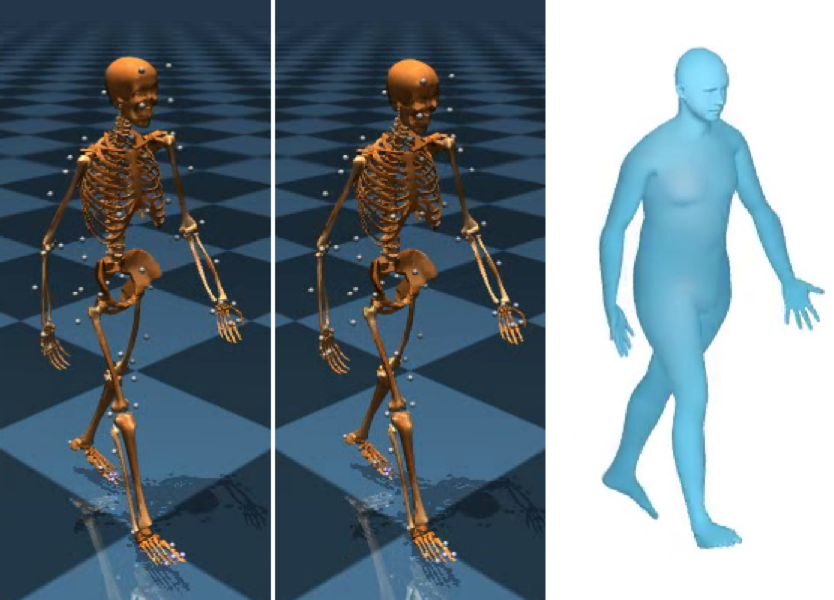}} &
            \fbox{\includegraphics[width=0.47\linewidth]{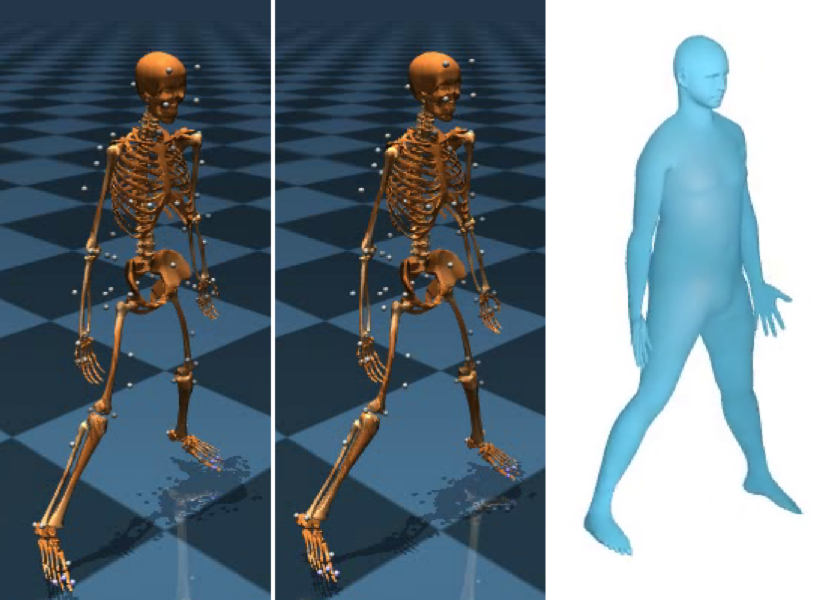}} \\[5pt]
            \fbox{\includegraphics[width=0.47\linewidth]{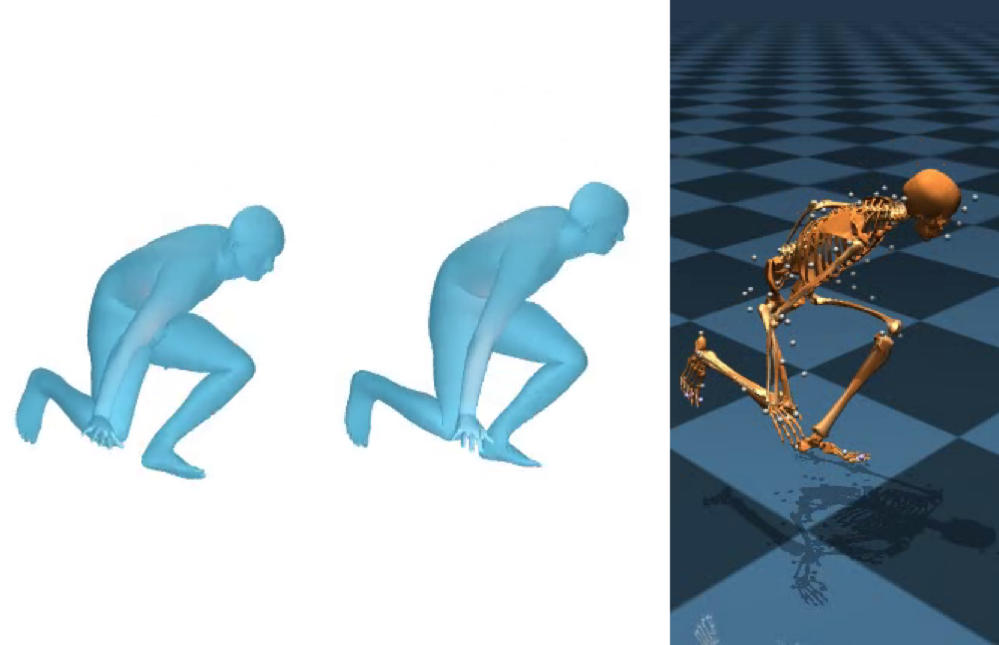}} &
            \fbox{\includegraphics[width=0.47\linewidth]{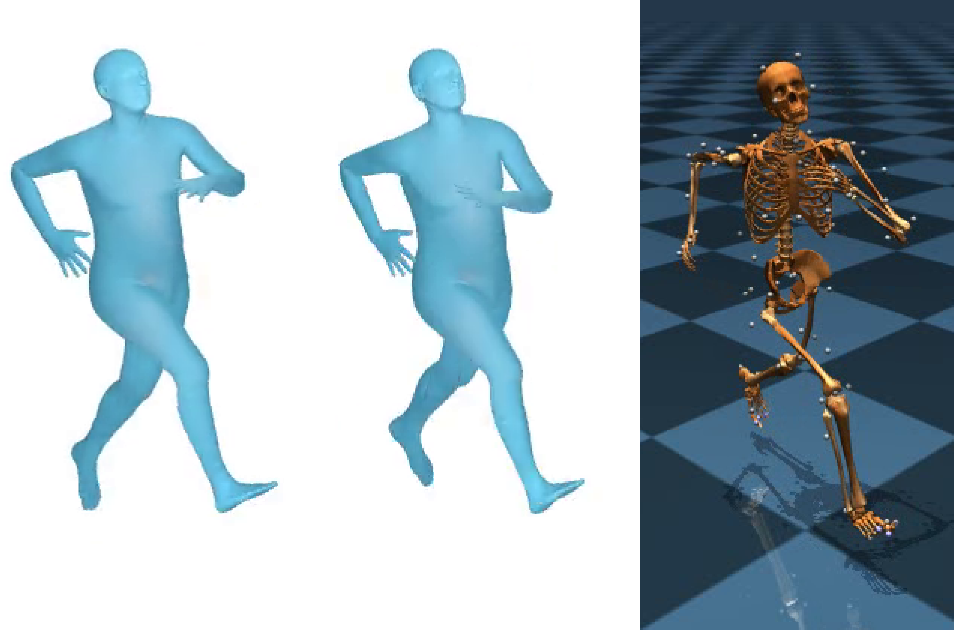}} \\
        \end{tabular}
        \end{minipage}
    \end{minipage}
    \caption{\textbf{Tokenizer training results.}
    \textbf{(a)} Codebook usage across the two datasets after training, showing a shared latent space. \textbf{(b)} Reconstruction examples on clinical (top row) and HumanML3D (bottom row) datasets. Within each panel: ground-truth pose (left), reconstructed pose (middle), and pose converted to the other format (right).}
    \label{fig:vqvae_plot}
\end{figure*}

We trained the VQVAE tokenizer with a codebook of 512 codewords, each a 512-dimensional vector. The model achieved low reconstruction error and high codebook usage (Table~\ref{table:vqvae_result}; \textit{Recon} is the mean absolute error of joint-angle reconstruction in degrees; \textit{Perplexity} measures codebook usage, with higher values indicating more uniform utilization and a maximum of 512; \textit{Conversion} reports the mean absolute error, in centimeters, between ground-truth and converted site locations across formats). For comparison, we trained another tokenizer on the clinical dataset alone, using a single encoder–decoder pair and no SMPL-related loss, showing training on both dataset achieved comparable reconstruction loss. The codebook was shared between the two datasets (Fig.~\ref{fig:vqvae_plot}a), confirming that the model learned format-independent representations and indicating that, despite differences in data format and source, the two datasets can be described with a common vocabulary of movements. Feeding trials through one encoder and reading out the other format's decoder produces a conversion between dataset formats (Fig.~\ref{fig:vqvae_plot}b; example videos are provided in the supplementary materials). We note that our tokenizer approach can be extended to additional body model-based motion datasets, provided that a correspondence or similarity metric between site locations can be defined (e.g. there is mapping between MoVi and SMPL-X keypoint locations \cite{ghorbani_soma_2021, pavlakos_smplx_2019}).

\begin{table}[ht]
\centering
\footnotesize
\setlength{\tabcolsep}{3pt}
\caption{Tokenizer training results.}
\begin{tabular}{lcccccc}
\toprule
Dataset & \multicolumn{3}{c}{Clinical Dataset} & \multicolumn{3}{c}{HumanML3D Dataset} \\
 & Recon & Perplexity & Conversion & Recon & Perplexity & Conversion \\
\midrule
Both dataset & 4.98 & 298.12 & 3.35 & 7.02 & 300.81 & 4.29 \\
Clinical only & 4.70 & 321.72 & - & - & - & - \\
\bottomrule
\end{tabular}
\label{table:vqvae_result}
\end{table}

\subsection{Language Model Performance}
We next trained BiomechGPT on multimodal language + motion inputs using our tokenized motion data, and evaluated the capacity of the model to perform a suite of 10 clinical motion assessment tasks: 6 classification and 4 regression.

\subsubsection{Classification Tasks}
\label{subsec:cls_task}
\begin{figure*}[!t]
    \centering

    \begin{minipage}[t]{0.75\textwidth}
        \vspace{7pt}
        \centering
        \begin{overpic}[width=\linewidth]{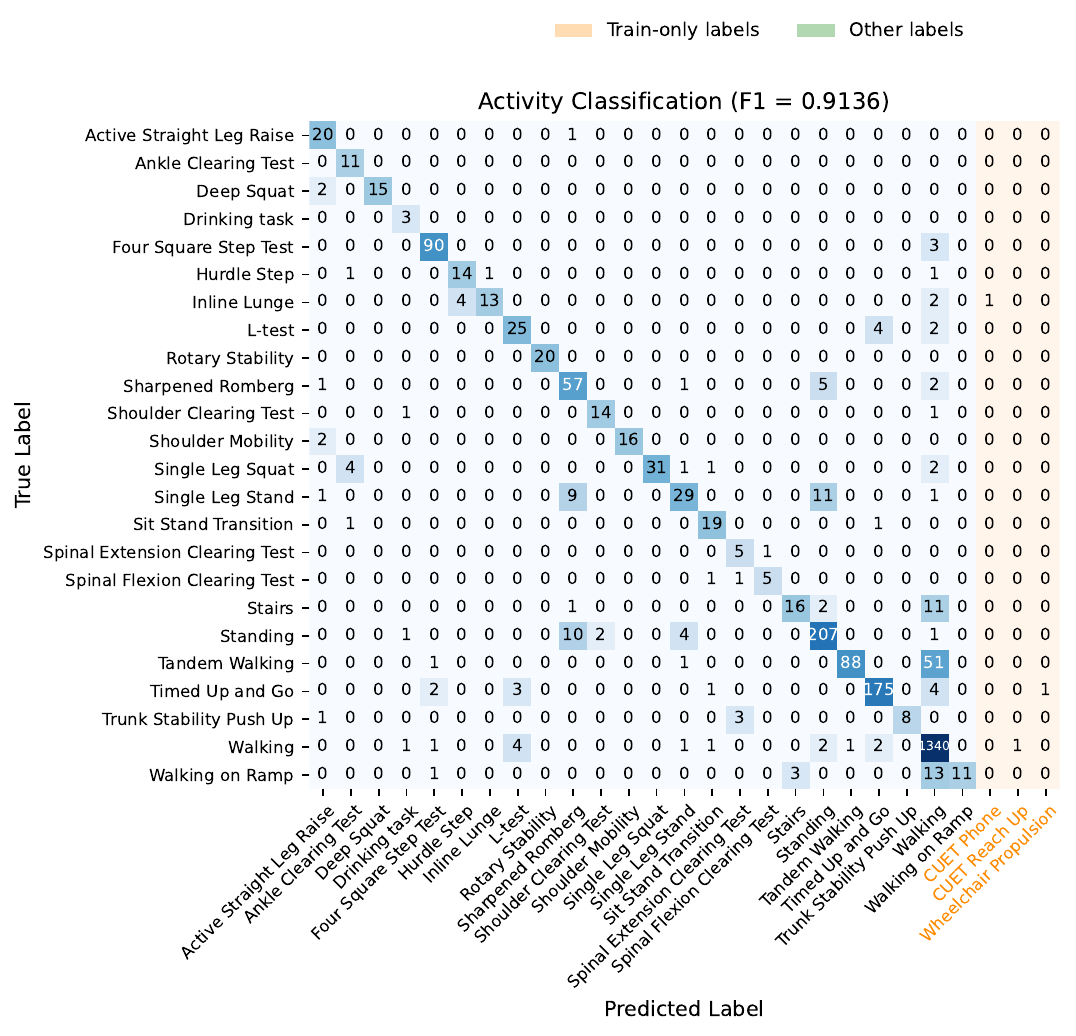}
            \put(3,87){\textbf{(a)}}
        \end{overpic}    \end{minipage}
    \hfill
    \begin{minipage}[t]{0.22\textwidth}
        \vspace{10pt}
        \centering

        \begin{overpic}[width=\linewidth]{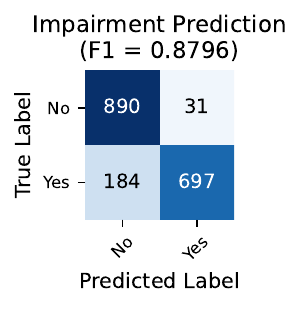}
            \put(-10,90){\textbf{(b)}}
        \end{overpic}
        \vspace{0pt}
        
        \begin{overpic}[width=\linewidth]{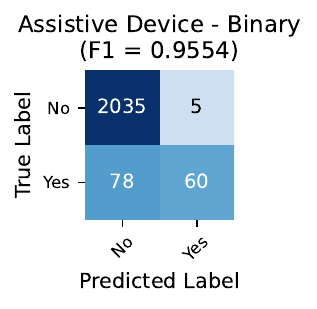}
            \put(-10,90){\textbf{(c)}}
        \end{overpic}
        \vspace{0pt}

        \begin{overpic}[width=\linewidth]{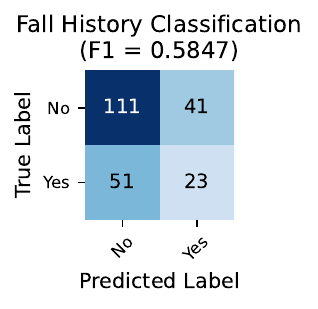}
            \put(-10,90){\textbf{(d)}}
        \end{overpic}
        \end{minipage}

    \begin{minipage}[t]{0.47\textwidth}
        \vspace{10pt}
        \centering
        \begin{overpic}[width=\linewidth]{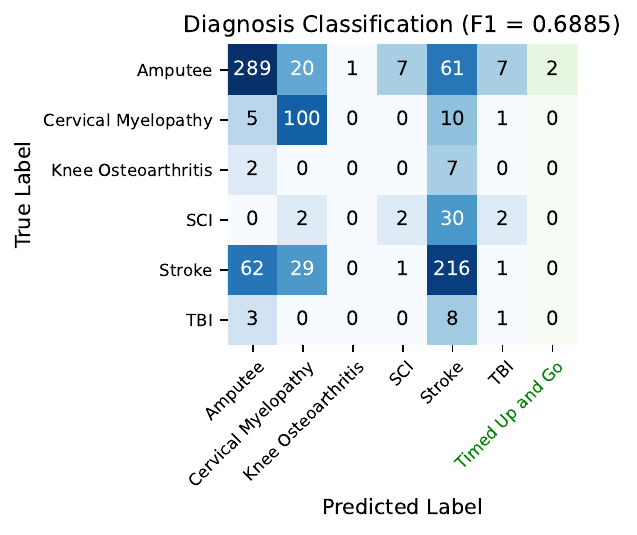}
            \put(3,80){\textbf{(e)}}
        \end{overpic}
        \end{minipage}
    \hfill
    \begin{minipage}[t]{0.47\textwidth}
        \vspace{10pt}
        \centering
        \begin{overpic}[width=\linewidth]{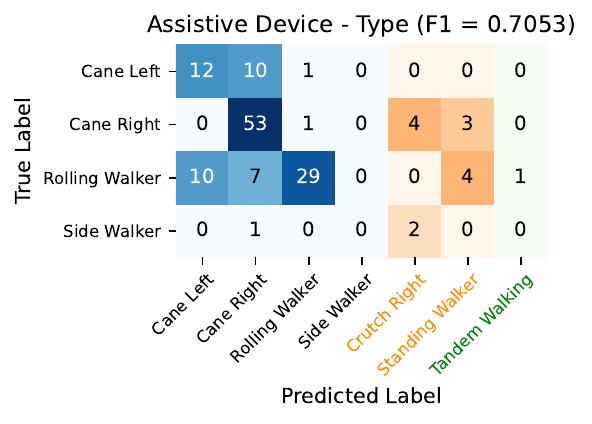}
            \put(3,67){\textbf{(f)}}
        \end{overpic}
        \end{minipage}
    \vspace{0pt}
    \caption{\textbf{Confusion matrices for classification tasks.} 
    \textbf{(a)} Activity classification. 
    \textbf{(b)} Binary classification of movement impairment presence. 
    \textbf{(c)} Binary classification of assistive device presence. 
    \textbf{(d)} Binary classification of fall history within the past 12 months among lower-limb prosthesis users. 
    \textbf{(e)} Diagnosis classification among participants with movement impairment. 
    \textbf{(f)} Assistive device type classification among participants using a device. 
    Orange indicates predictions that appear in the training set but not the evaluation set; green indicates predictions that do not answer the current task.}
    \label{fig:cls_matrix}
\end{figure*}

We first evaluated performance on each of  six binary and multiclass classification tasks for activity recognition, diagnosis, and assistive device usage. We computed performance on each task using the weighted F1 score, with example confusion matrices shown in Fig.~\ref{fig:cls_matrix}. Several tasks exhibit class imbalance, which is reflected in the misclassification patterns discussed below.

Activity classification achieves an F1 score of 0.91, which is notable given the range of actions and presence of highly similar postures in the dataset. Off-diagonal misclassifications occur primarily between closely related activities, such as walking / tandem walking / walking on a ramp, and standing / sharpened Romberg.

The model performed well on the binary tasks of impairment prediction and assistive device detection, achieving F1 scores of 0.88 and 0.96, respectively, although we note a tendency of the model to default to "no assistive device" due to class imbalance on this task.

The multiclass classification tasks of diagnosis type and assistive device type proved more challenging, with F1 scores of 0.69 and 0.71, respectively. Among participants with impairments, misclassifications were again affected by class imbalance, with errors concentrated in less represented diagnoses, including spinal cord injury (SCI), traumatic brain injury (TBI), and knee osteoarthritis. The performance of classifying assistive device type here may have been limited by the absence of wrist and forearm angles in the input, which are informative for detecting hand-held devices such as canes and walkers.

The most difficult task was inferring the fall history of lower-limb prosthesis users, a binary task with an F1 score of 0.58.  Lower-limb prosthesis users report relatively high fall rates, so the dataset was approximately balanced for this task, though relatively small owing to the limited number of lower-limb prosthesis users in the population. Poor performance on this task may simply reflect its difficulty: positive labels correspond to any user who has had a fall in the past year, which is hard to gauge from current motion data.

\subsubsection{Regression Tasks}
\label{subsec:reg_task}
\begin{figure*}[!ht]
    \centering
    \begin{overpic}[width=0.24\linewidth]{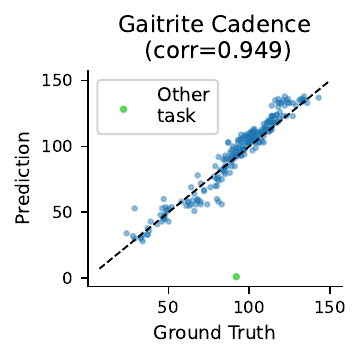}
        \put(2,92){\textbf{(a)}}
    \end{overpic}
    \begin{overpic}[width=0.24\linewidth]{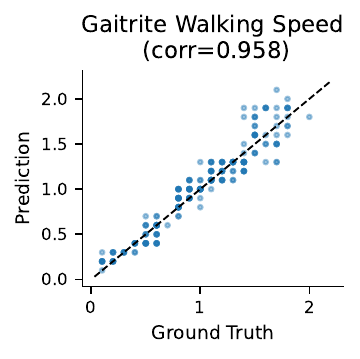}
        \put(2,92){\textbf{(b)}}
    \end{overpic}
    \begin{overpic}[width=0.24\linewidth]{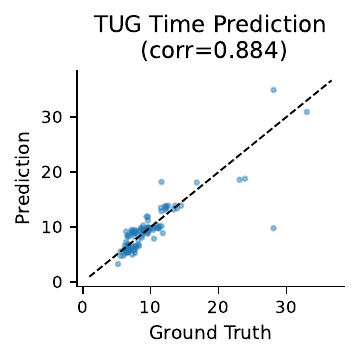}
        \put(2,92){\textbf{(c)}}
    \end{overpic}
    \begin{overpic}[width=0.24\linewidth]{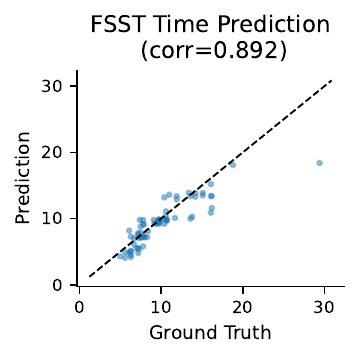}
        \put(2,92){\textbf{(d)}}
    \end{overpic}
    \caption{\textbf{Performance on regression tasks.} 
    \textbf{(a)} Walking cadence (steps/min), ground truth measured by GAITRite. 
    \textbf{(b)} Walking speed (m/s), ground truth measured by GAITRite. 
    \textbf{(c)} Time to complete the Timed-Up-and-Go (TUG) test. 
    \textbf{(d)} Time to complete the Four-Square-Step Test (FSST). Dashed line indicates $x=y$; green points indicate predictions that do not answer the current task.}
    \label{fig:reg_plot}
\end{figure*}

We next evaluated BiomechGPT on four regression tasks with numerical outputs: the cadence and average speed during walking on a GAITRite walkway, and the time to completion in the Timed-Up-and-Go (TUG) test and the Four Square Step Test (FSST), timed manually in the clinic. These external measurements provide ground-truth walking features that are independent of our motion-capture system. To evaluate model performance on each task, we extracted the numerical output from BiomechGPT responses to task prompts, and reported the Pearson correlation between ground truth and predicted values, with examples plotted in Fig~\ref{fig:reg_plot}.

Notably, the cross-entropy loss used in language modeling is not well-suited to numerical targets, so the model occasionally produces repeated digits or outliers, as reported in \cite{singh_2024_tokenization_counts, golkar_2024_xval, zausinger_2025_regress_loss} and shown in suppl. section~\ref{suppl:reg_sample}. To alleviate this effect, for each regression prediction we sampled the answer 5 times with temperature=0.5 and top\_p=0.5, and took the median of the 5 values as the final prediction.

BiomechGPT performed well on all four tasks, with Pearson's correlation values of 0.95 for cadence prediction, 0.96 for speed prediction, 0.88 for TUG time prediction, 0.89 for FSST time prediction. Thus, despite language models not being optimized for numerical tasks, we found that BiomechGPT could perform well on clinical regression tasks without additional alignment strategies. 

\subsubsection{Instruction-following ability}
\label{subsec:instruct_follow}
In both the classification and regression tasks, we observed a small number of trials where the model failed to follow instructions and produced an answer that did not correspond to the current task. These are indicated in green in the confusion matrices and regression plots: for example, the prediction of ``Timed Up and Go'' for diagnosis classification (Fig.~\ref{fig:cls_matrix}), and returning speed when asked about cadence (Fig.~\ref{fig:reg_plot}). Such failures are rare, and the average failure rate across all runs shown in Fig.~\ref{fig:model_scaling} (details in the following section) was 0.041\%, computed as the ratio of the number of failure cases to the total number of generated outputs. 

In addition, due to small sample sizes some classification labels occurred for only one or two patients; these were included in the training set but not in the test set. Occurrences of the model generating these labels on test data are marked in orange (Fig.~\ref{fig:cls_matrix}). Because these labels are a real part of the training set, we do not consider their generation to be a failure of instruction-following.

\subsection{Performance on single-camera video}
\label{subsec: pbl}
One-third of our clinical biomechanics dataset was captured from single-camera smartphone video \cite{peiffer_2025_pbl}, a fast and accessible way to collect movement data. BiomechGPT performs well on these single-camera data (suppl. Table~\ref{table:lm_mmc_pbl}, rows 2--4), showing that the model does not does not depend on data acquisition methods and can be used in more accessible settings.

\subsection{Scaling base model and dataset size}
\label{subsubsec: scaling}
Scaling model and dataset size is known to improve performance in general-purpose LLMs \cite{kaplan_2020_scalinglaws, hoffmann_2022_chinchilla}, here we test whether the same holds for clinical motion tasks. We trained BiomechGPT with four backbones of increasing capacity: Gemma-3 1B, Gemma-3 4B, MedGemma-1.5 (a 4B model further fine-tuned on medical text and images), and Gemma-3 12B. For each backbone, we also assessed the effect of training dataset size using two dataset variants: one with only the clinical dataset and clinical tasks, and one that added HumanML3D data and its associated tasks, which target general motion modeling rather than clinical objectives.

\begin{figure*}[!t]
    \centering
    \includegraphics[width=\linewidth]{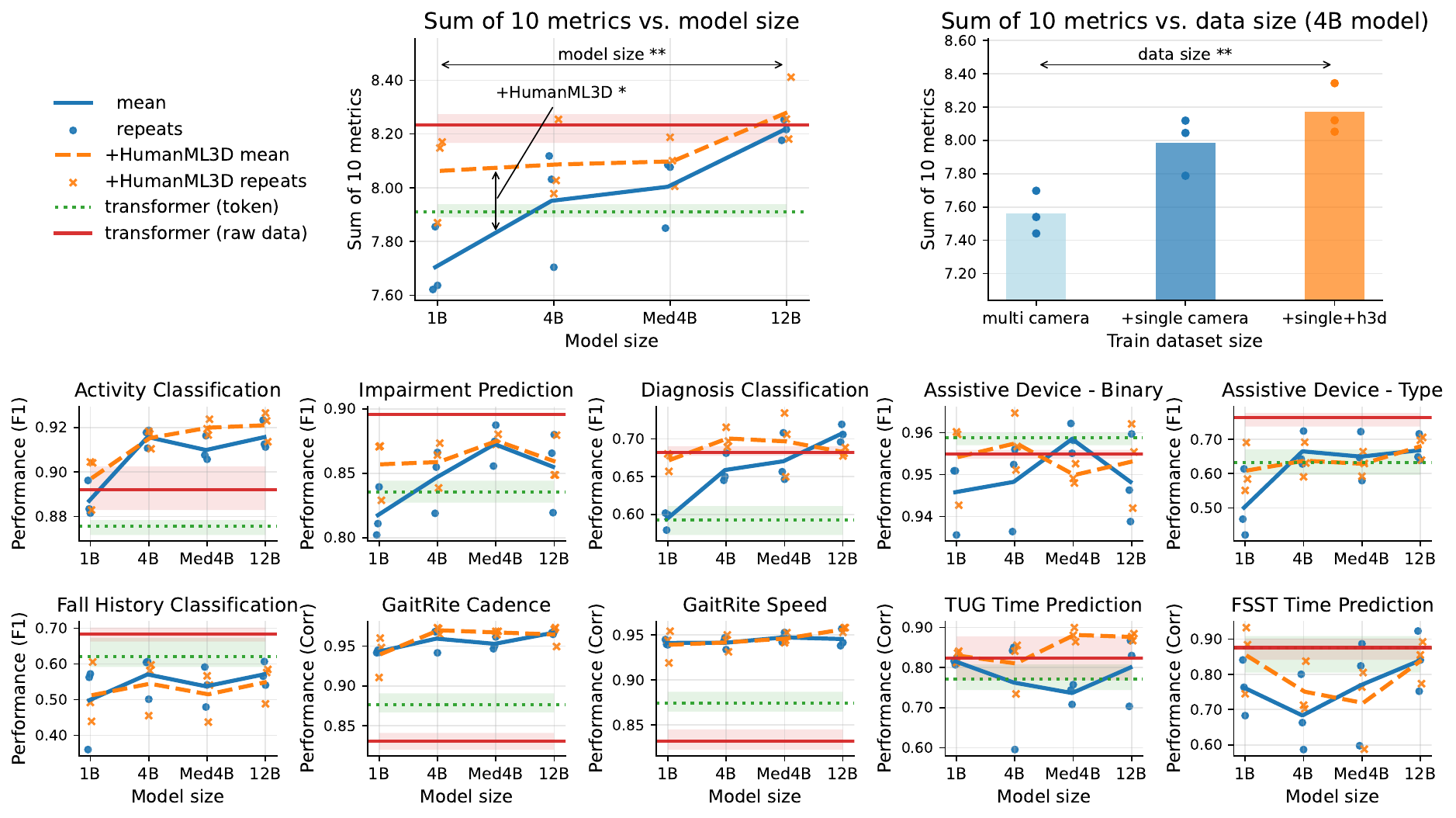}
\caption{\textbf{Language model performance improves with larger model and dataset sizes.} (Top left) Sum of the ten tasks as model size increases, trained with (orange, dashed) and without (blue, solid) added HumanML3D data; lines are means over repeats, scatters are repeats. Two-way ANOVA: significant effects of model size (p=0.003, **) and added data (p=0.011, *), no significant interaction. (Top right) 4B model as training data expands from multi-camera, to multi- plus single-camera, to also adding HumanML3D (one-way ANOVA, p=0.007, **); evaluated on the multi-camera test set only, while all other panels use the combined test set. (Remaining panels) Per-task performance versus model size. Green dotted and red solid lines mark the token-level and raw-data transformer baselines, shading shows max/min across three repeats.}
    \label{fig:model_scaling}
\end{figure*}

Fig.~\ref{fig:model_scaling} reports performance on each of the 10 clinical motion assessment tasks across all model variants and repeats (weighted F1 for classification, Pearson correlation for regression). Because per-task performance can be noisy for language models, we also report the sum across all 10 tasks as a summary statistic (fig.~\ref{fig:model_scaling} top left). A table of per-task results and chance values is reported in the supplement (suppl. section~\ref{suppl:lm_10task}, chance computed by randomly sampling training split answers as prediction). Although scaling effects vary across individual tasks, aggregate performance rises steadily with both model and dataset size, and a two-way ANOVA confirms that both factors are significant (model size: p=0.003; additional data: p=0.011). While interaction between model and dataset size effects is not significant, a trend emerges where the benefit of additional data is most pronounced at smaller model sizes, and the 1B model trained with HumanML3D data outperforms both the 4B and Med4B models trained without it. As model size grows, additional data continues to yield smaller but consistent gains. To look at data scaling more directly, in Fig.~\ref{fig:model_scaling} (top right) we fixed the backbone at 4B and expanded the training data in three stages: multi-camera clinical recordings only, then adding single-camera recordings, then also adding HumanML3D. Aggregate performance rises at each stage, and a one-way ANOVA across the three data levels confirms a significant effect of data size (p=0.007). To keep the comparison fair as the training data changes, this panel is evaluated on the multi-camera test set only.

We conducted two ablations on dataset composition within the clinical dataset (see suppl. section~\ref{suppl:2vs10}, \ref{suppl:mmcvspbl} for details). First, we verified that training on the full 10-task set does not hurts model performance on individual tasks, by training an additional set of models using only two core tasks: activity classification and impairment classification. A three-way ANOVA confirmed that model size (p\textless0.001) and dataset size (p=0.002) remained significant factors in the changes of model performance, while task set did not (p=0.598). Second, as our clinical dataset is drawn from two sources (multi- and single-camera recording), we verified that combining these sources improved performance on several tasks and left the rest unchanged. We therefore included all 10 tasks and both data sources in our main experiments.

Importantly, annotations of movement content on our clinical dataset were restricted to a fixed list of activity types; they therefore do not capture finer motion features that vary within each task, especially for participants with movement impairments. As a result, when prompted to describe a trial, a model trained on clinical data alone can only produce labels drawn from the fixed set. We observed that incorporating the HumanML3D dataset, with its rich natural-language descriptions of movements, enabled BiomechGPT to generate more detailed and naturalistic descriptions for clinical data as well (see supplementary section for example videos). This richer supervision may help the model develop a deeper representation of the underlying motion features.

Together, these findings indicate that scaling both model capacity and training data produces broad performance improvements on biomechanics-based clinical tasks, and that additional data remains beneficial even when drawn from partially or entirely different sources. This suggests that training on larger, more heterogeneous datasets and combining a wider range of tasks will continue to yield broad performance gains.


\subsection{Comparison to transformer models}
\label{subsec:baseline_compare}
We have shown that BiomechGPT can be used for motion interpretation tasks in a clinical dataset, where it provides a natural language interface to query both categorical and numerical aspects of human motion data. However, whether this interface shows any performance advantages over other, non-language-model-based methods for motion interpretation is unclear. To address this question, we compared BiomechGPT to two non-language transformer models using our 10 clinical tasks: we name these transformer-token, which consumes the same motion tokens as BiomechGPT and isolates the contribution of the language-model architecture, and transformer-raw, which consumes raw joint angles and serves as a strong motion analysis baseline with direct access to the continuous pose signal. Performance of these two models are shown as horizontal lines in Fig.~\ref{fig:model_scaling} and listed in suppl.~Table~\ref{table:lm_10task}.

We found that BiomechGPT outperforms transformer-token on most individual tasks, and on aggregate for most model variants. This indicates that the language-model architecture generally extracts more information from the tokenized representation than a non-language transformer with matched input.

However, the comparison against the stronger transformer-raw was more mixed: BiomechGPT matched or exceeded transformer-raw on Activity, Diagnosis, Cadence, Speed, and TUG (with +HumanML3D), while transformer-raw retained an advantage on Impaired, Assistive Type, and Fall History. As the model size grows, at the 12B scale, BiomechGPT matches transformer-raw in aggregate performance. Notably, all BiomechGPT models substantially outperformed both transformer variants on the GAITRite cadence and speed regression tasks, where transformer-token also surpassed transformer-raw; this suggests that tokenization itself provides a useful inductive bias for these tasks, despite the continuous nature of the target output itself.

\subsection{Effect of tokenization on language model performance}
\label{subsec:tokenization}
VQ-based tokenization is the most common strategy for integrating motion data into motion-language models~\cite{jiang_motiongpt_2023, guo_2023_momask, chen_2024_language_of_motion, qu_2024_llm_good_action_recognizers}, but quantizing continuous motion into a finite vocabulary inevitably loses information. \cite{zhu_2025_motiongpt3} found that in a diffusion framework, continuous embeddings from the MLD VAE \cite{chen_2023_mld} consistently outperform VQVAE tokens for motion generation tasks, while \cite{qu_2024_llm_good_action_recognizers} found that continuous embeddings did not outperform tokens for action recognition.

In our own experiments, the non-language transformer trained on raw joint angles (transformer-raw) outperformed the one trained on VQVAE tokens (transformer-token) by $+0.32$ in aggregated 10-task performance (bottom block of Table~\ref{table:tokenization_effect}; Figure~\ref{fig:model_scaling}, Sum of 10 panel), a measurable cost from discretization. This motivated us to perform an early exploration of how tokenization choices affect downstream performance. We keep VQ-based tokenization, the common choice across the works cited above, as our baseline, and test two directions for increasing tokenizer capacity.

We first asked whether higher codebook capacity helps (top block of Table~\ref{table:tokenization_effect}). Increasing the VQVAE codebook from $K=512$ to $K=2048$ reduced our tokenizer's reconstruction error from 4.98 to 4.52, and switching to a residual VQVAE ($K=2\times256$), which represents each time point as a sum of two tokens, reduced it further to 3.98. Downstream performance did not follow: both higher-capacity variants underperformed the $K=512$ baseline. Codebook capacity therefore does not appear to be the bottleneck.

We next replaced hard tokens with soft tokens, feeding continuous motion representations into the language model's embedding space rather than extending its vocabulary (middle block of Table~\ref{table:tokenization_effect}; see Methods). Against the hard-token baseline with matched training strategy, VQVAE-derived soft tokens improved task performance by $+0.49$ and whole-trial VAE soft tokens by $+0.43$, while the windowed VAE variant reduced it by $-0.17$. Reconstruction quality again failed to track downstream performance: the windowed VAE had the best reconstruction error (2.81, almost half of VQVAE's) yet performed worst on our 10 clinical interpretation tasks, while the whole-trial VAE had the worst reconstruction (5.75) but outperformed it.

Across the strategies we tested, reconstruction quality was a poor predictor of a language model's motion-understanding ability, and higher reconstruction quality did not recover the cost of tokenization. Improvements in tokenizer reconstruction therefore do not automatically translate into language-model gains, possibly because the two are optimized under different objectives, an effect also observed in vision-language modeling \cite{gu_2023_rethink_vq_tokenizer, wang_2025_endtoendvisiontokenizer, song_2026_dualtoken, qu_2025_tokenflow}. Better motion encoding~\cite{zhang_2026_encoder_free_motion} and motion-language alignment strategies to close this gap remain open directions for future work.

\begin{table*}[ht] 
\centering 
\footnotesize
\setlength{\tabcolsep}{3pt}
\caption{Effect of tokenization on downstream clinical task performance. All task models trained on the 10 clinical tasks. \textit{Clinical Recon.\ error}: joint-angle reconstruction MAE (degrees). \textit{Sum of 10}: aggregate score across the 10 tasks. \textit{Delta} is computed within each block relative to its first row.}
\begin{tabular}{l|cccccc}
\toprule
Tokenizer & Clinical Recon. error & Task model & Token type & Sum of 10 & Delta \\
\midrule
VQVAE (K=512) & 4.98 & Gemma 4B  & hard & 7.9512 & - \\
VQVAE (K=2048) & 4.52 & Gemma 4B  & hard & 7.8080 & -0.1432 \\
RVQVAE (K=2x256) & 3.98 & Gemma 4B  & hard & 7.8746 & -0.0766 \\
\midrule
VQVAE (K=512) & 4.98 & Gemma 4B  (answer only) & hard & 7.3517 & - \\
 & 4.98 & projector+Gemma 4B (answer only) & soft & 7.8445 & +0.4928 \\
VAE (D=512) & 5.75 & projector+Gemma 4B  (answer only) & soft & 7.7827 & +0.4310 \\
 & 2.81 & projector+Gemma 4B (answer only) & soft & 7.1860 & -0.1657 \\
\midrule
VQVAE (K=512) & 4.98 & non-language transformer & hard &  7.9111 & - \\
Raw data & 0.00 & non-language transformer & - &  8.2344 &  +0.3233 \\
\bottomrule
\end{tabular}
\label{table:tokenization_effect}
\end{table*}

\section{Discussion}
Our work shows that recent advances in motion–language modeling can be effectively translated to clinical movement analysis, and that clinical motion understanding in turn provides a precise, measurement-grounded benchmark for evaluating these models. We introduced BiomechGPT, a motion-language model for clinical movement understanding, together with a cross-format shared representation motion tokenizer that embeds biomechanical and SMPL motion into a shared latent space without requiring paired data. To our knowledge, this is the first work to evaluate motion-language models on a diverse set of clinical movement QA tasks. A single instruction-tuned model handled multiple clinical motion classification and regression tasks, outperforming a matched-input transformer baseline and matching a stronger transformer trained on raw kinematic signals in aggregate at the 12B scale, with substantial advantages on activity recognition, walking cadence, and speed prediction, where tokenization appears to provide a useful inductive bias. These results held for data captured with both multi-camera systems and single smartphone video, showing the approach does not depend on how the movement was recorded and can be used in more accessible settings.

Unlike open-ended generation or captioning, these clinical tasks have objective ground-truth labels and continuous measurements, allowing precise, fine-grained evaluation of motion–language models. We observed clear scaling effects, with performance rising with both larger model and dataset size. Our dataset is large by clinical biomechanics standards but small relative to typical language-model corpora, making it important to leverage large public motion datasets, since labeled clinical data is scarce. Incorporating general motion–language data improved clinical task performance, indicating such datasets carry useful information about human movement, and our tokenizer offers an efficient way to integrate heterogeneous formats. Directly fitting pose parameters per trial reaches lower cross-format error but is far slower ($\sim$3.5 min/trial), whereas the tokenizer can be reused across tasks. Our results suggest that training on larger, more heterogeneous datasets \cite{lin_2024_motionx, fan_2025_motion_million} and a wider range of tasks should keep yielding gains, though an open question is whether such large general-purpose data will begin to swamp out the clinical signal. Nonetheless, the competitive performance of a single LLM unifying classification, regression, and description under one interface, and its continued improvement with more data, supports our broader hypothesis: a large biomechanics-language model is a promising generalist framework for movement analysis, and clinical motion understanding is a meaningful application and evaluation setting for the motion–language modeling community.

\subsection{Limitations}
Our tokenization study showed that lower reconstruction error did not translate into better downstream task performance, highlighting tokenization as an important factor in motion–language modeling and suggesting that reconstruction objectives are not well aligned with downstream language-model training. While BiomechGPT performance was competitive overall, non-language transformer models that could take continuously valued joint angle input retained a per-task advantage on tasks where fine-grained continuous kinematic detail appears critical. Improvements in tokenizer design, motion-language alignment, or further scaling are promising directions for closing this gap.

At the representation level, our tokenizer discards root translation and orientation, which removes information about global movement dynamics; future work will explore ways to incorporate root information while keeping joint angles as the primary representation, consistent with how clinicians interpret movement data. We zero out several joints due to tracking reliability, which limited the learning of hand movement in particular. Including arm and hand biomechanics is an important future direction, as their high range of motion and open-ended nature make them difficult to quantify, presenting a critical gap in computational rehabilitation assessment; tokenized hand representations have shown promise in this regard~\cite{wang_2024_motiongpt2}.

In the future, the clinical dataset could be expanded with denser annotations, such as additional clinical information, action segmentation, and patient recovery trajectories, as well as multi-turn prompts to support open-ended conversation. Another potential improvement is to combine BiomechGPT with the agentic tools we developed for analyzing biomechanics data \cite{cotton_2026_biomechagent} and to further train it using reinforcement learning with verifiable rewards (RLVR) \cite{lambert_2025_rlvf}. Several tasks were affected by class imbalance arising from the small participant population; addressing this through targeted data collection, class-balanced sampling \cite{Chawla_2002_smote}, or synthetic motion augmentation \cite{adeli_2025_gaitgen} is a promising direction. Fall-history classification remained the hardest task, limited both by its inherent difficulty and by the small number of lower-limb prosthesis users. Regression with a cross-entropy loss produces occasional outliers that are only partially alleviated by sampling-based aggregation, so integrating regression-aware losses or numerical tokenization schemes is a natural next step \cite{zausinger_2025_regress_loss, singh_2024_tokenization_counts}. Finally, like many language models, BiomechGPT is prone to hallucinations and to producing errors confidently; it cannot explain its outputs, and it does not provide uncertainty estimates such as differential diagnoses or confidence scores, all of which are important for trustworthy clinical use—limitations. These limitations could be addressed through RLVR and reasoning for motion \cite{wang_2026_unimo, wang_2025_agir, chen_2026_biogaitvlm} to provide interpretable, evidence-grounded predictions.

\section{Conclusion}
We introduced BiomechGPT, a multimodal motion–language model for clinical movement understanding, together with a cross-format tokenizer that embeds motion from different body models into a shared latent space without suing paired data. A single instruction-tuned language model handled diverse clinical classification and regression tasks, with competitive performance that improved as we scaled up both model and dataset size. These results show that clinical motion understanding is both a precise benchmark for evaluating motion–language models and a meaningful application of them, giving clinicians a natural language interface to biomechanical data. We hope this supports further development of motion–language models for rehabilitation and movement science.



\section*{Acknowledgment}

Research reported in this publication was supported in part by the Eunice Kennedy Shriver National Institute Of Child Health \& Human Development of the National Institutes of Health under Award Number R01HD114776 (RJC), a Pew Biomedical Scholar award (AK), and a McKnight Scholar award (AK).


\FloatBarrier  
\clearpage
\ifisr
    {\small\printbibliography}
\fi

\clearpage
\section{Supplementary Material}

\subsection{Supplemental video}
\label{suppl:video}
We provide supplemental videos to illustrate our results. First, we provide videos showing reconstruction and data format conversion following tokenizer training between clinical and HumanML3D dataset. The three panels on the videos from left to right are ground truth (GT), reconstructed and converted motion. Second, we provide videos showing GT motion trials from clinical dataset and motion descriptions generated for them. We show that by training on natural language descriptions for HumanML3D dataset, BiomechGPT is able to generate natural description for clinical dataset, where such description is not available. This would be useful to capture variations in performing the same task, especially for people with movement impairment. The videos and descriptions are generated with the same tokenizer and language model runs as in the example task plots in main content. As we remove root translation and rotation, some poses may look different. We also note that the tokenizer is optimized for clinical data reconstruction, thus does not have optimal performance for HumanML3D objectives.

\subsection{Biomechanical reconstruction method}
\label{suppl:biom_method}
We process our data using a method similar to \cite{cotton_differentiable_2025}, we refer readers to it for full details. Here we provide a brief summary of our biomechanical reconstruction process. However, due to clinical privacy protection, we are not able to make this dataset public.

We acquired synchronized multiview video while participants performed the different activities. Each frame is cropped around the participant, and a pose-estimation network (MeTRAbs-ACAE \cite{sárándi_2022_metrabs}) extracts 87 anatomically meaningful 2-D key-points per camera view. We used a biomechanical model that was modified from LocoMujoco \cite{alhafez_2023_locomujoco}, which was derived from the OpenSim model \cite{hamner_muscle_2010}. We modified this to include a 3-degree-of-freedom neck joint. Instead of trying to triangulate those points directly, our method learns an implicit neural function of time that outputs joint angle values. Those angles drive the skeleton model which has eight learnable scale factors and small per-marker offsets to account for different body scales. Forward kinematics inside the simulator then turn the angles into coherent 3-D virtual-marker trajectories, which is compared to the direct estimation and the error between them is optimized for. The fitting process results in a trajectory for each trial $\mathbf \tau = \left\{ \boldsymbol \beta, \mathbf  q_0, \dot {\mathbf  q}_0, \mathbf q_1, \dot  {\mathbf  q}_1 ... \mathbf q_T, \dot  {\mathbf  q}_{T} \right\}$. $\beta$ corresponds to the body shape. $\mathbf q$ is a 41-element vector where the first 3 elements are the global position in Euclidean space, the next 4 correspond to a quaternion representation of the pelvis orientation in 3-D space, and the remaining 34 elements are the body joint angles. $\dot {\mathbf q}$ is the change in these parameters, which is a 40-element vector as the change in the quaternion is represented as a 3-element rotational vector.

\subsection{Example question-answer pairs for each task}
\label{suppl:all_qa}
The full list of tasks and an example question-answer pair for each is provided in Suppl. Table~\ref{table:full_task}. For each task we have multiple question templates, and one of them is randomly selected for each motion trial when creating the dataset for training. For HumanML3D tasks, we use question prompts from MotionGPT \cite{jiang_motiongpt_2023} as a reference.

\begin{table*}[ht] 
\centering 
\footnotesize
\setlength{\tabcolsep}{3pt}
\caption{Full list of tasks and example question-answer pairs.}
\vspace{-6pt}
    \begin{tabular}{p{3.5cm} | p{9cm} | p{3.5cm}}
    \toprule
    Task & Question & Answer \\
    \midrule
    Clinical dataset: &  &  \\
    \midrule
    activity classification & [motion token] What type of activity is this person doing? & [activity]\\
    \midrule
    impaired classification & [motion token] Does someone who moves like this likely have a movement impairment? & [movement impairment] \\
    \midrule
    diagnosis classification & [motion token] What is the most likely diagnosis for their gait impairment? & [diagnosis] \\
    \midrule
    assistive device presence & [motion token] Does the way this person moves imply the use of an assistive device? & [assistive device binary] \\
    \midrule
    assistive device type & [motion token] What type of assistive device is the person using in this motion? & [assistive device type] \\
    \midrule
    fall history classification & Has the individual shown in [motion token] fallen at least once in the past 12 months? (Yes/No) & [fall binary] \\
    \midrule
    gaitrite cadence & [motion token] What is the cadence of this walking in steps/min? & [cadence] steps/min \\
    \midrule
    gaitrite walking speed & [motion token] What is the speed of this walking in m/s? & [speed] m/s \\
    \midrule
    TUG time prediction & [motion token] How long did it take this person to complete the timed-up-and-go task? & [TUG time]\\
    \midrule
    FSST time prediction & [motion token] How long did it take this person to complete the four-square-step-test? & [FSST time]\\
    \midrule
    HumanML3D dataset: &  & \\
    \midrule
    description & Describe the motion represented by [motion token] using plain English. & [range description] \\
    \midrule
    range time & [motion token] Find the start and end time range of a person doing '[range description]' in the given motion sequence in second. & [range start] [range end] \\
    \midrule
    range duration & [motion token] Tell me the duration in seconds that the person performs '[range description]'. & [range duration] \\
    \midrule
    range description & [motion token] What happened in this motion sequence between [range start] and [range end] seconds? & [range description] \\
    \bottomrule
    \end{tabular}
    \label{table:full_task}
\end{table*}

\subsection{Regression results without sampling}
\label{suppl:reg_sample}
\begin{figure*}[!ht]
    \centering
    \begin{overpic}[width=0.24\linewidth]{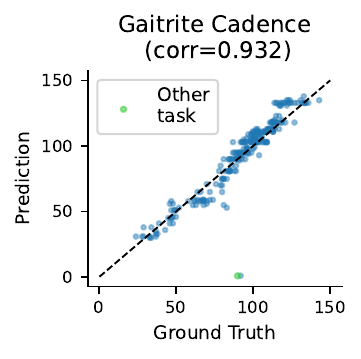}
        \put(2,92){\textbf{(a)}}
    \end{overpic}
    \begin{overpic}[width=0.24\linewidth]{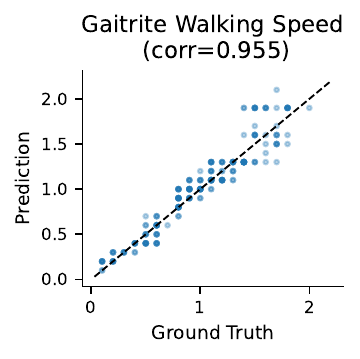}
        \put(2,92){\textbf{(b)}}
    \end{overpic}
    \begin{overpic}[width=0.24\linewidth]{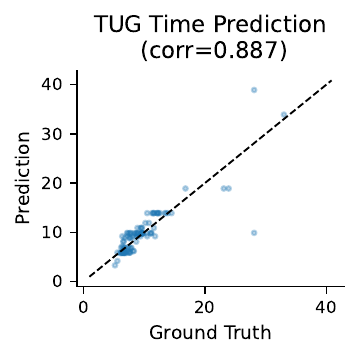}
        \put(2,92){\textbf{(c)}}
    \end{overpic}
    \begin{overpic}[width=0.24\linewidth]{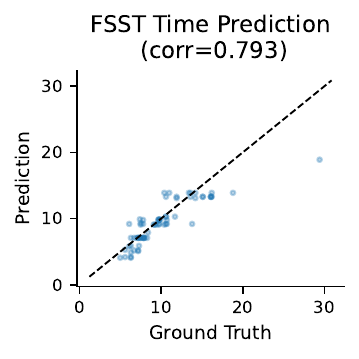}
        \put(2,92){\textbf{(d)}}
    \end{overpic}
    \caption{\textbf{Regression task without sampling.} Same notation as in fig~\ref{fig:reg_plot}, but without sampling 5 times and take the median for each prediction. Noticeable horizontal lines are shown in plots, which means the model tends to generate the same numarical value. Dashed line indicates $x=y$; green points indicate predictions that do not answer the current task.}
    \label{fig:reg_plot_no_sample}
\end{figure*}

We found that when generating numerical outputs for regression tasks, the language model tends to generate the same answer repeatedly, resulting in visible horizontal lines in the plot. Therefore, in the main text we report regression results by sampling five times for each prediction with temperature=0.5 and top\_p=0.5, and take the median of the 5 answers. This method reduced the horizontal line effect a lot. Here we provide regression task plots (suppl. Fig~\ref{fig:reg_plot_no_sample}) generated with the same BiomechGPT checkpoint as in main text, but without sampling, showing horizontal lines.

\subsection{Per-task performances with model scaling}
\label{suppl:lm_10task}
Here we provide the pre-task results for model scaling in supplementary Table~\ref{table:lm_10task} for Fig~\ref{fig:model_scaling} top left and per-task breakdown, and per-task results for data size scaling in Table~\ref{table:lm_bar_plot} for Fig~\ref{fig:model_scaling} top right. We also provide the per-task performance of the two non-language transformer baseline and chance value, evaluated on the combined data from multi- and single-camera sources. Chance is computed by randomly sampling answers in the training set for each task.

\begin{table}[ht] 
\centering 
\footnotesize
\setlength{\tabcolsep}{3pt}
\caption{Per-task performance on 10 tasks, tested on combined data with scaling model size and training data size.}
\resizebox{1.0\textwidth}{!}{%
\begin{tabular}{l|cccccc|cccc|c}
\toprule
Model & Activity & Impaired Y/N & Diagnosis & Assist Y/N & Assist Type & Fall Y/N & Cadence & Speed & TUG & FSST & Sum of 10 \\
\midrule
1B & 0.8870 & 0.8175 & 0.5935 & 0.9458 & 0.5008 & 0.4982 & 0.9428 & 0.9410 & 0.8154 & 0.7624 & 7.7044 \\
1B+h3d & 0.8972 & 0.8569 & 0.6716 & 0.9542 & 0.6086 & 0.5121 & 0.9396 & 0.9389 & 0.8297 & 0.8539 & 8.0626 \\
\midrule
4B & 0.9157 & 0.8466 & 0.6587 & 0.9483 & 0.6648 & 0.5707 & 0.9593 & 0.9412 & 0.7626 & 0.6833 & 7.9512 \\
4B+h3d & 0.9153 & 0.8587 & \underline{0.7005} & 0.9575 & 0.6376 & 0.5446 & \textbf{0.9698} & 0.9416 & 0.8102 & 0.7508 & 8.0865 \\
\midrule
Med4B & 0.9099 & 0.8726 & 0.6703 & \underline{0.9585} & 0.6493 & 0.5373 & 0.9528 & \underline{0.9471} & 0.7364 & 0.7694 & 8.0037 \\
Med4B+h3d & \underline{0.9199} & \underline{0.8752} & 0.6967 & 0.9499 & 0.6284 & 0.5148 & \underline{0.9670} & 0.9459 & \textbf{0.8813} & 0.7189 & 8.0980 \\
\midrule
12B & 0.9157 & 0.8550 & \textbf{0.7070} & 0.9483 & 0.6678 & 0.5710 & 0.9666 & 0.9454 & 0.8001 & 0.8386 & 8.2154 \\
12B+h3d & \textbf{0.9210} & 0.8590 & 0.6822 & 0.9532 & \underline{0.6804} & 0.5496 & 0.9645 & \textbf{0.9562} & \underline{0.8761} & 0.8404 & \textbf{8.2826} \\
\midrule
transformer (token) & 0.8756 & 0.8353 & 0.5923 & \textbf{0.9588} & 0.6318 & \underline{0.6210} & 0.8769 & 0.8743 & 0.7709 & \underline{0.8742} & 7.9109 \\
transformer (raw data) & 0.8920 & \textbf{0.8959} & 0.6824 & 0.9549 & \textbf{0.7632} & \textbf{0.6834} & 0.8306 & 0.8321 & 0.8242 & \textbf{0.8757} & \underline{8.2342} \\
\midrule
\shortstack{chance$\pm$\\std} & \shortstack{0.3350$\pm$\\ 0.0063} & \shortstack{0.4995$\pm$\\0.0119} &  \shortstack{0.3403$\pm$\\0.0149} &  \shortstack{0.8558$\pm$\\0.0044} &  \shortstack{0.3381$\pm$\\0.0392} &  \shortstack{0.4787$\pm$\\0.0333} & \shortstack{0.0030$\pm$\\0.0624} & \shortstack{-0.0015$\pm$\\0.0664} & \shortstack{0.0006$\pm$\\0.1017} & \shortstack{-0.0026$\pm$\\0.1220} &  \\
\bottomrule
\end{tabular}
}
\label{table:lm_10task}
\end{table}

\begin{table*}[ht] 
\centering 
\footnotesize
\setlength{\tabcolsep}{3pt}
\caption{Per-task performance of 4B model on 10 tasks, tested on multi-camera data with scaling training data size.}
\resizebox{1.0\textwidth}{!}{%
\begin{tabular}{l|cccccc|cccc|c}
\toprule
Model & Activity & Impaired Y/N & Diagnosis & Assist Y/N & Assist Type & Fall Y/N & Cadence & Speed & TUG & FSST & Sum of 10 \\
\midrule
4B+Multi-camera only & 0.9152 & 0.8563 & 0.6510 & 0.9534 & 0.6058 & 0.5547 & 0.9580 & 0.9329 & 0.5901 & 0.5421 & 7.5596 \\
4B+Combined & 0.9177 & 0.8621 & 0.6564 & 0.9431 & 0.6590 & 0.5563 & 0.9593 & 0.9412 & 0.7934 & 0.6950 & 7.9835 \\
4B+Combined+h3d & 0.9263 & 0.8689 & 0.6998 & 0.9540 & 0.6198 & 0.5392 & 0.9698 & 0.9416 & 0.8774 & 0.7750 & 8.1718 \\
\bottomrule
\end{tabular}
}
\label{table:lm_bar_plot}
\end{table*}

\subsection{Model performance on two- vs. 10-task set}
\label{suppl:2vs10}
We retrained the same BiomechGPT variants on a two-task set containing only the two core tasks: activity classification and impairment classification. We compared these with the performance of the two tasks from the full 10-task training. A three-way ANOVA showed that model size and dataset size remained significant factors, while the number of training tasks did not (Suppl. Table~\ref{tab:lm_sum2task}; model size: p\textless0.001; additional data: p=0.002; task set: p=0.598), indicating that adding auxiliary tasks does not degrade performance on the core tasks. We therefore included all 10 tasks in our main experiments.

\begin{table*}[h]
\centering
\caption{Model performance on two- vs. 10-task set, tested on combined data}
\label{tab:lm_sum2task}
\setlength{\tabcolsep}{6pt}
\resizebox{1.0\textwidth}{!}{%
\begin{tabular}{llcccc}
\toprule
Model size & Task set & HumanML3D & Activity classification & Impairment classification & Sum of 2\\
\midrule
1B    & 2  & $-$ & 0.8706 & 0.8573 & 1.7279 \\
      & 2  & $+$ & 0.8775 & 0.8508 & 1.7283 \\
      & 10 & $-$ & 0.8870 & 0.8175 & 1.7045 \\
      & 10 & $+$ & 0.8972 & 0.8569 & 1.7541 \\
\midrule
4B    & 2  & $-$ & 0.9079 & 0.8644 & 1.7723 \\
      & 2  & $+$ & 0.9118 & 0.8567 & 1.7685 \\
      & 10 & $-$ & 0.9157 & 0.8466 & 1.7623 \\
      & 10 & $+$ & 0.9153 & 0.8587 & 1.7740 \\
\midrule
Med4B & 2  & $-$ & 0.9003 & 0.8398 & 1.7401 \\
       & 2  & $+$ & 0.9081 & 0.8567 & 1.7648 \\
       & 10 & $-$ & 0.9099 & 0.8726 & 1.7825 \\
       & 10 & $+$ & 0.9199 & 0.8752 & 1.7951 \\
\midrule
12B   & 2  & $-$ & 0.9131 & 0.8674 & 1.7805 \\
      & 2  & $+$ & 0.9236 & 0.8943 & 1.8179 \\
      & 10 & $-$ & 0.9157 & 0.8550 & 1.7707 \\
      & 10 & $+$ & 0.9210 & 0.8590 & 1.7800 \\
\bottomrule
\end{tabular}
}
\end{table*}

\subsection{Model performance on different data sources}
\label{suppl:mmcvspbl}
The clinical dataset is drawn from two sources: multi-camera~\cite{cotton_differentiable_2025} and single-camera~\cite{peiffer_2025_pbl} recordings. For the 10 clinical tasks, we compared BiomechGPT with 4B base model size trained on the multi-camera source alone with ones trained on both sources (Suppl. Table~\ref{table:lm_mmc_pbl}). BiomechGPT performed well on both data sources. Additionally, the combined-source model improved performance on three tasks strongly and left the rest roughly unchanged, suggesting that additional training data is beneficial even when drawn from a partially distinct distribution.

\begin{table*}[ht] 
\centering 
\footnotesize
\setlength{\tabcolsep}{3pt}
\caption{4B model performance by train and test data sources.}
\resizebox{1.0\textwidth}{!}{%
\begin{tabular}{ll|cccccc|cccc}
\toprule
Train data & Test data & Activity & Impaired & Diagnosis & Assist Y/N & Assist Type & Fall & Cadence & Speed & TUG & FSST \\
\midrule
Multi-camera only & Multi-camera only &0.9152 & 0.8563 & 0.6510 & 0.9534 & 0.6058 & 0.5547 & 0.9580 & 0.9329 & 0.5901 & 0.5421 \\
Combined & Multi-camera only & 0.9177 & 0.8621 & 0.6564 & 0.9431 & 0.6590 & 0.5563 & 0.9593 & 0.9412 & 0.7934 & 0.6950 \\
Combined & Single-camera only & 0.9231 & 0.7900 & 0.7189 & 0.9652 & 0.8096 & 0.6112 & - & - & 0.7081 & 0.7430 \\
Combined & Combined & 0.9157 & 0.8466 & 0.6587 & 0.9483 & 0.6648 & 0.5707 & 0.9593 & 0.9412 & 0.7626 & 0.6833 \\
\bottomrule
\end{tabular}
}
\begin{tablenotes}
\item *There is no trial with cadence and speed measured for single-camera dataset.
\end{tablenotes}
\label{table:lm_mmc_pbl}
\end{table*}

\subsection{Tokenizer reconstruction performance with input length}
\label{suppl:input_len}
Following prior motion-tokenizer work, both the VQVAE and VAE are trained on fixed-length windows of 64 and applied to full trials at inference. We report reconstruction error on the clinical dataset (Suppl. Table~\ref{table:tokenizer_input_len}) under two encoding schemes: encoding the whole trial in a single pass, or encoding it as length-64 windows whose embeddings are then concatenated. Decoding is always done on the full-length embeddings. The VQVAE generalizes well to longer inputs, whereas the VAE encoder degrades substantially. This makes it possible to obtain VAE embeddings of differing quality (measured by reconstruction loss) and compare their downstream performance in BiomechGPT.

\begin{table}[ht] 
\centering 
\footnotesize
\setlength{\tabcolsep}{3pt}
\caption{Tokenizer reconstruction error (MAE) generalizing to input trial length for clinical dataset.}
    \begin{tabular}{l|cc}
    \toprule
    Tokenizer & Encode by full trial & Encode by window \\
    \midrule
    VQVAE (K=512) & 4.98 & 5.09\\
    \midrule
    VAE (D=512) & 5.75 & 2.81\\
    \bottomrule
    \end{tabular}
    \label{table:tokenizer_input_len}
\end{table}

\subsection{Additional ablation on tokenizer}
Upon surveying the literature on VQVAE-based motion models, we noticed that the VQVAE tokenizer model code in these works \cite{zhang_t2mgpt_2023, jiang_motiongpt_2023} is borrowed from Jukebox \cite{dhariwal_2020_jukebox} and Bailando \cite{siyao_2022_bailando}. The VQVAE encoder and decoder in these works are all composed of several residual network modules, each containing several residual convolutional blocks. In the two original works, these convolutional layers use increasing dilation in the encoder and decreasing dilation in the decoder. In the motion VQVAE models, however, both the encoder and decoder use decreasing dilation, and this change is not explained in the literature. We therefore conducted preliminary experiments on the dilation ordering in the VQVAE tokenizer. We report reconstruction error on the clinical and HumanML3D datasets, along with the aggregated downstream performance on the 10 clinical tasks using the 4B BiomechGPT. Arrows indicate the direction of dilation change across conv layers (encoder–decoder, $\uparrow$ for increasing, $\downarrow$ for decreasing). We chose the dilation setting that achieves decent reconstruction error on both datasets while giving slightly better downstream clinical task performance. These experiments again support our observation that improvements in reconstruction quality do not translate to language model performance. However, given the variability in language model performance, these values should be interpreted as indicative trends rather than precise comparisons.

\begin{table}[ht] 
\centering 
\footnotesize
\setlength{\tabcolsep}{3pt}
\caption{Influence of tokenizer dilation on reconstruction error (MAE) and BiomechGPT performance.}
    \begin{tabular}{c|ccc}
    \toprule
    Dilation & \shortstack{Clinical\\recon. error} & \shortstack{HumanML3D\\recon. error} & \shortstack{BiomechGPT\\performance} \\
    \midrule
    \shortstack{$\uparrow - \downarrow$\\(original VQVAE)}& 4.58 & 7.61 & 7.9350 \\
    \midrule
    \shortstack{$\downarrow - \downarrow$\\(previous motion-VQVAE)} & 5.28 & 7.05 & 7.7496 \\
    \midrule
    \shortstack{$\uparrow - \uparrow$\\(our model)} & 4.98 & 7.04 & 7.9512\\
    \midrule
    $\downarrow - \uparrow$ & 5.30 & 6.24 & 7.9129\\
    \bottomrule
    \end{tabular}
    \label{table:tokenizer_dilation}
\end{table}

\end{document}